\documentclass[lettersize,journal]{IEEEtran}
\usepackage{amsmath,amsfonts}
\usepackage{algorithm,algpseudocode}

\usepackage[caption=false,font=normalsize,labelfont=sf,textfont=sf]{subfig}
\usepackage{textcomp}
\usepackage{stfloats}
\usepackage{multirow}
\usepackage{url}
\usepackage{adjustbox}
\usepackage{verbatim}
\usepackage{graphicx, multicol}
\usepackage{cite}
\usepackage{array}
\usepackage{epstopdf}
\usepackage{color}
\usepackage{enumerate}
\usepackage{float}
\usepackage{booktabs}
\usepackage{siunitx}
\usepackage{subfig}
\usepackage{hyperref}
\usepackage{arydshln}
\usepackage{tikz}
\usepackage{balance} 
\usepackage{pifont}
\usepackage{wrapfig}

\makeatletter
\newcommand{\closetagspacing}{\renewcommand{\tag@spacing}{0pt}}
\makeatother
\algrenewcommand\algorithmicindent{0.6em}%
\hypersetup{
	colorlinks=False,
	linkcolor=cyan,
	filecolor=blue,
	urlcolor=red,
	citecolor=green,
}
\newcommand\highlightReference[1]{%
  \expandafter\newcommand\csname highlightReference-#1\endcsname{}%
}
\let\oldbibitem\bibitem
\def\bibitem#1 #2\par{%
  \expandafter\ifx\csname highlightReference-#1\endcsname\relax
    \oldbibitem{#1}#2\par
  \else
    \oldbibitem{#1}\highlight{#2}\par
  \fi
}

\newcommand\highlight[1]{\textcolor{red}{#1}}

\begin{document}
\title{CognitionCapturerPro: Towards High-Fidelity Visual Decoding from EEG/MEG via Multi-modal Information and Asymmetric Alignment}

\author{Kaifan Zhang,
        Lihuo He,~\IEEEmembership{Member,~IEEE},
        Junjie Ke,
        Yuqi Ji,
        Lukun Wu,
        Lizi Wang,~\IEEEmembership{Member,~IEEE},
        Xinbo Gao,~\IEEEmembership{Fellow,~IEEE}
\thanks{Kaifan Zhang, Lihuo He, Junjie Ke, Yuqi Ji, Lukun Wu and Xinbo Gao are with the Visual Information Processing Laboratory, School of Electronic Engineering, Xidian University, Xi'an 710071, China. (\emph{Corresponding author: Lihuo He}, e-mail: lhhe@mail.xidian.edu.cn.)}
\thanks{Lizhi Wang is with the School of Artificial Intelligence, Beijing Normal University, Beijing 100875, China}
}



\maketitle

\begin{abstract}
Brain decoding has attracted growing attention in recent years, with most existing studies foucsing on reconstructing visual stimuli with high fidelity. However, achieving accurate reconstruction of original visual stimuli from neural signals remains a substantial challenge. Neuroscience research identifies two key bottlenecks in visual decoding. First, when visual stimuli are transformed into neural signals by the visual system, partial information is inevitably lost, a phenomenon referred to as fidelity loss. Second, the brain's dynamic perceptual, cognitive, and associative processes introduce subjective biases, leading to representation shift. Under conditions of limited neural data, the resulting misalignment between neural representations and image features becomes a major obstacle, causing models to overfit and generalize poor.

To address these challenges, this paper proposes \textit{CognitionCapturerPro}, a framework that integrates EEG signals with multi-modal data including images, text, depth maps, and edge information through collaborative training. An uncertainty-weighted similarity scoring mechanism is first introduced to dynamically quantify the fidelity relationships between brain signals and other modalities. A fusion encoder is then designed to effectively integrate modality-specific and shared information. In the alignment phase, a simplified alignment module is employed to enhance reconstruction quality while accelerating training. For the reconstruction stage, a pre-trained diffusion model is incorporated to generate visual stimuli with coherent semantic and structural properties. Experiments conducted on the THINGS-EEG dataset demonstrates that compared to CognitionCapturer, our method improves Top-1 and Top-5 retrieval accuracy by 25.9\% and 10.6\%, respectively. The results provide a robust technical pathway for advancing neural decoding and brain-computer interfaces (BCIs) toword real-world applications such as medical rehabilitation and assistive technologies. The code will be made publicly available upon publication.
\end{abstract}

\begin{IEEEkeywords}
Brain-computer interface (BCI), contrastive learning, electroencephalography (EEG), magnetoencephalography (MEG), image recognition.
\end{IEEEkeywords}

\section{Introduction}

\IEEEPARstart
Humans can recognize objects in complex scenes within milliseconds\cite{dicarlo2007untangling}. This astonishing ability of ``what you see is what you know" has inspired researchers to propose the inverse problem: can visual experiences be directly ``read" from neural signals? In recent years, researchers have attempted to decode visual information from functional magnetic resonance imaging (fMRI), magnetoencephalography (MEG), and electroencephalography (EEG), opening new avenues for brain-computer interfaces and the study of human visual mechanisms\cite{benchetritbrain,chen2024visual,du2023decoding,takagi2023high,zhang2025neurobridge}. However, while fMRI and MEG are constrained by low temporal resolution or high costs \cite{allen2022massive, cichy2014resolving}, EEG’s portability, affordability, and millisecond-level resolution make it ideal for real-world settings \cite{spampinato2017deep, edelman2024non}.

\begin{figure}[t]
    \centering
    \includegraphics[width=1\linewidth]{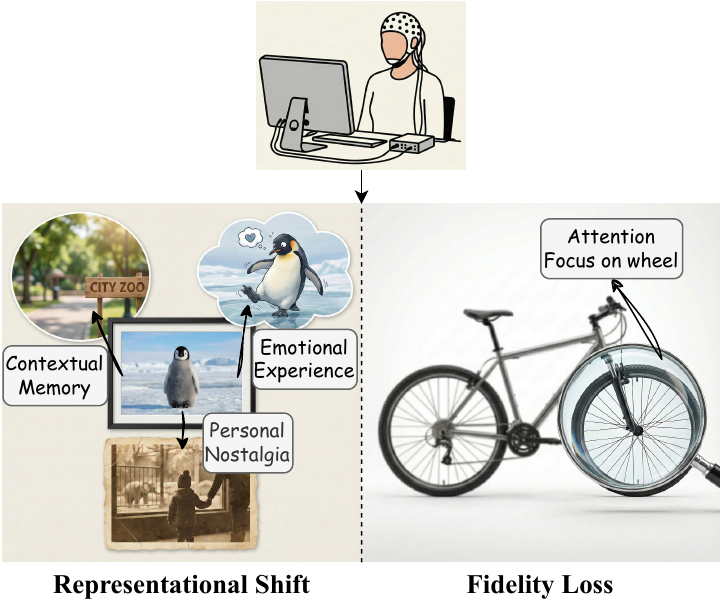}
    \caption{Core challenges in EEG-based visual decoding. We categorize the misalignment between brain signals and visual stimuli into two aspects: (Left) \textbf{Representational Shift}, where the brain's associative mechanisms introduce non-visual semantics that deviate from the pixel-level content; and (Right) \textbf{Fidelity Loss}, where selective attention and perceptual uncertainty result in incomplete or localized neural capture of the original image.}
    \label{fig:intro}
\end{figure}

Early studies have achieved impressive classification results and successfully generated saliency maps using EEG signals\cite{palazzo2020decoding}. However, the block-wise experimental design they employed, which grouped images of the same category into the same block, led to classification results relying more on temporal correlations at the block level rather than stimulus-related activity\cite{li2020perils}. It was not until recent years that a study significantly expanded this landscape by compiling a large and diverse EEG dataset\cite{gifford2022large}. this dataset contains 16,740 image stimuli, covering 1,854 concepts, and employs Rapid Serial Visual Presentation (RSVP) paradigm\cite{keysers2001speed}, avoiding the inflated accuracy caused by block designs. Based on this dataset, researchers have proposed various brain decoding methods aimed at aligning the representations of EEG signals with visual stimuli to achieve retrieval and reconstruction of the original visual stimuli\cite{du2023decoding,li2024visual,song2024decoding,song2025recognizing,zhang2025cognitioncapturer,zhang2025neurobridge,wu2025shrinking,wu2025bridging}. However, key challenges remain in accounting for the systematic differences between brain signals and visual stimuli. 

Neural signals are not simple mirrors of visual stimuli. During the transformation from visual stimuli to neural representation, information undergoes complex filtering and reorganization, which may lead to a significant misalignment between brain signals and the original images. As illustrated in Figure \ref{fig:intro}, we propose that this misalignment could be attributed to two major challenges:

First, what we term \textbf{representational shift}: the human brain's natural associative mechanisms activate semantic networks beyond the visual content itself when processing visual stimuli (e.g., seeing a penguin may associate with ice, Antarctica, or childhood zoo visits). The introduction of such non-visual semantic information causes the information contained in the brain signal to deviate from the visual features of the image. Second, what we term \textbf{fidelity loss}: limited by the observer's attentional mechanisms, human perception of images is often local and selective (e.g., focusing only on bicycle's wheel rather than the entire frame), or attention may shift, resulting in the visual information captured by the brain itself being incomplete and uncertain. These two challenges together constitute the core obstacles in decoding visual stimuli from EEG.


While progress has been made, existing methods have not fully overcome these two challenges simultaneously. One group of methods focuses on semantic alignment but assumes the neural signal is a complete representation of the stimulus. Consequently, these approaches neglect fidelity loss and remain limited in recovering fine visual details\cite{zhang2025cognitioncapturer,song2025recognizing,song2024decoding,wei2024mb2c}. Another group, designed to handle fidelity loss by simulating perceptual uncertainty, typically operates within a narrowly visual framework and misses the broader semantic associations of representational shift\cite{wu2025bridging,wu2025shrinking,zhang2025neurobridge}. Consequently, an integrated approach effectively combining both perspectives has yet to be realized.


In our conference work CognitionCapturer\cite{zhang2025cognitioncapturer}, we addressed the representational shift problem—where the brain’s associative mechanisms introduce non-visual semantics—by proposing a modal expansion strategy. This strategy extended original image-EEG pairs into multimodal sets (including image, text, and depth) to capture neural semantics beyond the visual stimulus. However, this framework did not explicitly account for fidelity loss, the inherent loss of information during neural processing, and its architecture left room for streamlining. To this end, we present CognitionCapturerPro, which incorporates several improvements to resolve these issues:

\begin{itemize}
\item \textbf{Uncertainty-Weighted Masking (UM):} which addresses fidelity loss by dynamically quantifies the brain-image fidelity relationship by combining human visual system-inspired spatially variable blurring with uncertainty-weighted similarity scoring. 

\item \textbf{Fusion Encoder:} which integrates modality-private and shared information using cross-modal attention and random modality dropout.

\item \textbf{Shared-Trunk \& Heads Alignment (STH-Align):} which obviates the need for a complex diffusion prior by incorporating a simplified MLP-based shared trunk, simultaneously receiving embeddings from all modalities and outputting aligned features, simplifying the architecture while improving alignment accuracy.
\end{itemize}

Our experimental results show that CognitionCapturerPro achieves state-of-the-art performance in zero-shot EEG-driven image retrieval and reconstruction. Representational similarity analysis shows clear clustering of object categories in the embedding space. Comprehensive neural analysis further validates the consistency of this framework with classic findings in cognitive neuroscience. Compared to the conference version, this paper adds experiments with MEG data, ablation studies for each module, and neuroscience-oriented experiments, further confirming the biological plausibility and generalizability of the framework.

\section{Related Work} \label{sec:related work}

\subsection{Brain Decoding}
Brain decoding refers to the process of interpreting brain signals to infer human perceptual and cognitive states. In recent years, this field has witnessed significant progress, particularly in applications such as motor imagery decoding\cite{aflalo2015decoding}, visual decoding\cite{du2023decoding,ren2021reconstructing,xia2024dream,li2024visual,song2025recognizing,zhang2025cognitioncapturer,zhang2025neurobridge,wu2025shrinking,wu2025bridging,benchetritbrain,song2024decoding}, text decoding\cite{duan2023dewave}, emotion decoding\cite{li2022eeg}, and the diagnosis of neurological disorders\cite{vicchietti2023computational}.

Visual decoding primarily involves two tasks: brain-to-image retrieval and reconstruction. A common foundational paradigm aligns the encoded representations of EEG/MEG signals with the embedding space of contrastive vision-language pre-trained models like CLIP~\cite{radford2021learning}. However, directly aligning these representations fails to account for the inherent modality gap between brain signals and visual stimuli, which often leads to overfitting on training data and poor generalization to new samples. Consequently, researchers have been refining this paradigm by incorporating insights from neuroscience, brain data structure, and deep learning to address these limitations. In terms of task specialization, some methods focus solely on the retrieval task~\cite{du2023decoding,li2024visual,song2025recognizing,song2024decoding,zhang2025neurobridge,wu2025shrinking}, while others handle both reconstruction and retrieval~\cite{zhang2025cognitioncapturer,li2024visual,benchetritbrain}.

\begin{figure*}[t]
    \centering
    \includegraphics[width=1\linewidth]{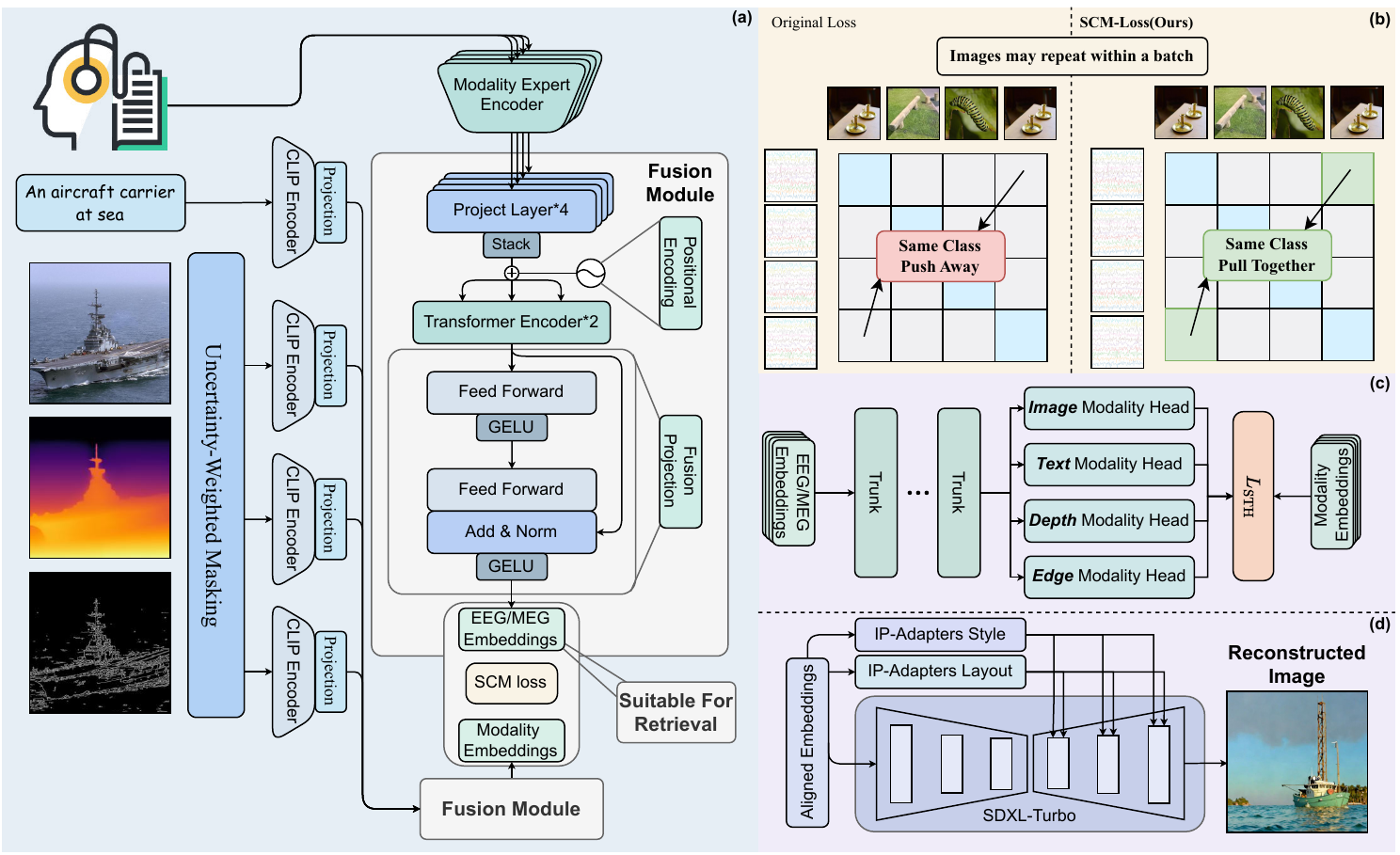}
    \caption{Overview of CognitionCapturerPro: Figure (a) depicts the encoder training process, where multimodal data such as EEG and images are processed by independent encoders, integrated through a fusion encoder, and constrained by an improved contrastive loss. (b) depicts the SCM-Loss, which filters positive pairs using semantic labels and similarity to address one-to-many mappings. (c) depicts the STH-Align structure, which maps embeddings from various modalities into a unified image space via a shared backbone and multimodal projection heads. (d) depicts how the aligned multimodal embeddings are conditionally injected through the SDXL-Turbo model with a multi-branch IP-Adapter to reconstruct semantically consistent and high-fidelity images.}
    \label{fig:framework}
\end{figure*}

\subsection{Cross-modal Contrastive Learning}

Recently, contrastive learning has achieved remarkable success across several domains~\cite{chen2020simple, gao2021simcse, you2020graph}. Building on this trend, multi-modal contrastive learning has emerged to map diverse modalities into a shared semantic space. A landmark advancement is the development of vision-language models, which exhibit exceptional zero-shot capabilities and generalization~\cite{stevens2024bioclip, wang2022medclip}. Inspired by this, researchers have extended the paradigm to various modality combinations~\cite{xue2023ulip, guzhov2022audioclip}.

In visual neural decoding, however, a notable ``dense-to-sparse" mismatch exists between dense visual features and sparse, noise-embedded brain representations. This discrepancy, compounded by inherent cognitive noise, hinders traditional alignment from capturing deep semantic correlations. To enhance decoding robustness, recent works have integrated uncertainty modeling into cross-modal contrastive learning frameworks to quantify data noise and dynamically adjust the learning focus~\cite{wu2025bridging, zhang2025neurobridge, wu2025shrinking}. However, purely visual uncertainty modeling neglects the broader semantic associations in the brain-to-visual transition.

\subsection{Generative Visual Decoding}   

In visual neural decoding, fMRI-based studies have evolved from early structural and texture modeling to high-fidelity reconstruction leveraging advanced generative models \cite{takagi2023high, kay2008identifying, scotti2024mindeye2}. In contrast, while EEG-based research has sought to emulate the success of fMRI, it remains fundamentally constrained by low signal-to-noise ratios. Consequently, current EEG methodologies have moved toward complex, multi-stage ``retrieval-then-reconstruction" strategies rather than a unified generative pipeline \cite{bai2024dreamdiffusion,xia2024dream,zhang2025cognitioncapturer,li2024visual}.

These approaches generally follow two technical routes: fine-tuning generative models to enhance reconstruction quality \cite{bai2024dreamdiffusion}, or aligning retrieval-optimized EEG embeddings with the CLIP latent space to facilitate high-fidelity synthesis \cite{xia2024dream,zhang2025cognitioncapturer,li2024visual}. While fine-tuning improves alignment, it often induces mode collapse and confines outputs to the training distribution. Conversely, mapping the EEG space to CLIP preserves generative diversity and fidelity by leveraging frozen foundation models. Additionally, recent research has begun to standardize evaluation metrics, providing more consistent benchmarks for assessing generative visual decoding \cite{xia2025multigranular}. 

\section{Method}

\subsection{Overview}


We formulate brain decoding as a dual task comprising \textbf{Retrieval} and \textbf{Reconstruction}. Let $\mathcal{E} = \{ \mathbf{E}^{(i)} \in \mathbb{R}^{C \times T} \}_{i=1}^{N}$ denote an EEG corpus with $C$ channels and $T$ time steps, and $\mathcal{M} = \{ M_m^{(i)} \}_{m \in \{ \text{img, text, depth, edge} \}}$ for $i=1,\dots,N$ represent the four synchronously captured modalities (image, text, depth map, and edge map), forming one-to-many pairs with each EEG segment.

\textbf{Retrieval Objective:}
Given an EEG sample $\mathbf{E}$, retrieve the most probable image $M_{\text{img}}$ by maximizing the posterior probability $P(M_{\text{img}} \mid \mathbf{E})$. \textbf{Reconstruction Objective:}
Given $\mathbf{E}$, synthesize a high-fidelity image $\hat{M}_{\text{img}}$ that minimizes the perceptual distance to the ground-truth image $M_{\text{img}}$.

Our proposed framework consists of five core components:

\begin{enumerate}
    \item \textbf{Uncertainty-Weighted Masking:}
    Inspired by human foveated vision, this module applies spatially-varying blur based on model confidence, thereby adaptively suppressing noisy alignments during training.

    \item \textbf{Modality Expert Encoders:}
    Four dedicated encoders transform raw EEG signals into modality-specific embeddings $\mathbf{z}_m$, where $m \in \{\text{eeg, txt, depth, edge}\}$, to preserve the distinctive characteristics of each modality.

    \item \textbf{Fusion Encoder:}
    A cross-modal Transformer model integrates all modality embeddings with learnable modality tokens, ultimately producing a unified representation $\mathbf{z}_f$ that captures complementary information.

    \item \textbf{Shared Backbone and Head Alignment:}
    A lightweight shared network, followed by modality-specific projection heads, maps both the individual embeddings and the fused embedding into a common image embedding space, enabling robust cross-modal alignment.

    \item \textbf{SDXL-Turbo + IP-Adapter Generation:}
    The aligned embeddings condition a fast diffusion model via multiple IP-Adapters (corresponding to image, depth, edge, etc.) to generate high-fidelity visual stimuli that are semantically consistent with the brain signals.
\end{enumerate}

This architecture mitigates the modality gap through aligned representations while safeguarding intra-modal details. It ensures consistent retrieval and reconstruction performance with limited brain data. Figure~\ref{fig:framework} illustrates the detailed architecture.

\subsection{Uncertainty-Weighted Masking (\texorpdfstring{UM}{UM})}

Inspired by \cite{wu2025bridging}, this method builds upon the human foveal vision mechanism. The core intuition is to dynamically adjust the learning difficulty: increasing peripheral blur for \textbf{``easy"} samples to prevent overfitting, while reducing it for \textbf{``hard"} samples to encourage focus on essential central features.

\subsubsection{Fovea-Inspired Spatially-Variant Blurring}

Given an input image \(\mathbf{I}\), we first generate a weight mask \(\mathbf{M}_{\text{fovea}}\) that gradually varies from the center to the edges. This mask has a value of 1 at the image center and decays exponentially to 0 towards the edges:
\begin{equation}
\mathbf{M}_{\text{fovea}}(i, j) = r_{\text{edge}} + (r_{\text{centre}} - r_{\text{edge}}) \cdot \exp \left( -\lambda \frac{d_{ij}}{d_{\max}} \right)
\end{equation}
where \((i, j)\) are pixel coordinates, \(d_{ij}\) denotes the distance from the pixel to the image center, \(d_{\max}\) is the maximum distance from the center to a corner, \(r_{\text{centre}}\) is the center weight (typically set to \(1 - c\)), \(r_{\text{edge}}\) is the edge weight (typically set to \(0\)), and \(\lambda\) is a parameter controlling the decay rate.

The final foveated image is then obtained by blending the clear original image with a blurred image using the weight mask:
\begin{equation}
\widetilde{\mathbf{I}} = \mathbf{M}_{\text{fovea}} \odot \mathbf{I} + (\mathbf{1} - \mathbf{M}_{\text{fovea}}) \odot \mathbf{I}
\end{equation}
where \(\odot\) denotes element-wise multiplication. This operation keeps the central region of the image clear while applying blur effects to the peripheral regions.

\subsubsection{Uncertainty-Driven Dynamic Selection}

We dynamically modulate the blur intensity $\sigma$ based on the model's current alignment performance to adaptively adjust task difficulty. For a mini-batch of $B$ samples, we first compute the similarity score $s_i$ between each image and its corresponding EEG embedding. To maintain a stable estimation of performance, we employ a memory bank $\mathbf{S}$ to store historical scores, updated via an exponential moving average (EMA):
\begin{equation}
\hat{s}_i = \gamma \cdot s_i + (1 - \gamma) \cdot \mathbf{S}_{\text{old}}[i], \quad \mathbf{S}[i] \leftarrow \hat{s}_i
\end{equation}
where $\gamma$ is the smoothing coefficient. Assuming the scores in $\mathbf{S}$ follow a Gaussian distribution, we calculate the mean $\hat{\mu}$ and standard deviation $\hat{\sigma}$ to establish a confidence interval $[\hat{\mu} - z\hat{\sigma}, \hat{\mu} + z\hat{\sigma}]$, where $z$ controls the sensitivity of the uncertainty threshold.

The blur intensity $\sigma$ for each sample is then adaptively selected based on its smoothed score $\hat{s}_i$:
\begin{equation}
\sigma = 
\begin{cases}
\sigma_0 - c, & \text{if } \hat{s}_i < \hat{\mu} - z \cdot \hat{\sigma} \quad \text{(Hard samples)} \\
\sigma_0 + c, & \text{if } \hat{s}_i > \hat{\mu} + z \cdot \hat{\sigma} \quad \text{(Easy samples)} \\
\sigma_0, & \text{otherwise} \quad \text{(Baseline)}
\end{cases}
\end{equation}
where $\sigma_0$ denotes the baseline blur radius, and $c$ is a controllable step size that modulates the adjustment margin. This uncertainty-driven mechanism ensures consistent generalization across diverse neural-visual pairs.

\subsection{Modality Expert Encoder}
The Modality Expert Encoder maps raw EEG segments into high-level semantic embeddings. Four parameter-isolated parallel branches are dedicated to image, text, depth, and edge modalities, guaranteeing that modality-private information is not diluted. The entire flow is written as
\begin{equation}
\mathbf{e} = \mathrm{Proj} \circ \mathrm{TS\text{-}Conv} \circ \mathrm{Attn}(\mathbf{E}), \qquad \mathbf{E} \in \mathbb{R}^{C \times T}.
\end{equation}
$\circ$ denotes function composition, indicating a right-to-left execution order. The $\mathrm{Attn}$ module identifies salient spatio-temporal regions by applying self-attention sequentially across channel and temporal axes. $\mathrm{TS\text{-}Conv}$ utilizes a depthwise separable structure, where 1D causal convolutions capture temporal dynamics and channel-wise convolutions aggregate spatial cross-lead dependencies. A two-layer MLP ($\mathrm{Proj}$) then maps these features into a unified hidden dimension to yield the modality-specific embedding $\mathbf{e}$.

Notably, the one-to-many mapping inherent in EEG-image datasets creates a training paradox for INFO-NCE loss: it simultaneously pulls and pushes samples of the same semantic category. To resolve this conflicting supervision, we introduce \textbf{Similarity-Category Masked Loss} (SCM-Loss). Its core idea is to involve only semantically identical and most similar sample pairs in the positive probability computation, while masking out all others.

Given EEG embeddings $\mathbf{E} \in \mathbb{R}^{B \times d}$ and labels $\mathbf{y} \in \mathbb{N}^B$, we first compute the temperature-scaled similarity matrix $\mathbf{S} = \mathbf{E}\mathbf{E}^\top / \tau$. To resolve semantic conflicts, we define a masked probability matrix $\mathbf{M}$ as:
\begin{equation}
\begin{aligned}
    M_{ij} &= \frac{\exp(S_{ij} \cdot m_{ij})}{\sum_{l=1}^B \exp(S_{il} \cdot m_{il})}, \\
    m_{ij} &= 
    \begin{cases} 
    1, & \text{if } y_i = y_j \text{ and } j \in \text{top-}k(S_{i, \cdot}) \\
    0, & \text{otherwise}
    \end{cases}
\end{aligned}
\end{equation}
where $m_{ij}$ denotes the similarity-category mask. The final SCM-Loss is formulated as the negative log-likelihood of the self-alignment:
\begin{equation}
    \mathcal{L}_{\text{SCM}} = -\frac{1}{B} \sum_{i=1}^B \log M_{ii}
\end{equation}

This approach ensures that only samples sharing the same semantic category and exhibiting top-$k$ similarity are treated as positive pairs, while same-class but dissimilar samples are suppressed as negatives, and cross-class samples are completely excluded from positive consideration. Consequently, the EEG embedding space evolves to maximize both intra-class compactness and inter-class separability, benefiting downstream cross-modal alignment and reconstruction tasks.

\subsection{Fusion Encoder}

The Fusion Encoder $\mathcal{F}_{\text{FE}}$ integrates four modality-specific EEG embeddings $\{\mathbf{z}_m\}$ into a unified representation $\mathbf{z}_{\text{fus}} \in \mathbb{R}^{1024}$. To effectively capture both private and synergistic information, the fusion process is conducted in three stages:

\subsubsection{Input Projection and Tokenization} 

To align the diverse modalities into a shared hidden space ($d=1024$), each embedding $\mathbf{z}_m$ is first mapped via a linear projection with LayerNorm and GELU activation: $\tilde{\mathbf{z}}_m = \mathrm{LN}(\mathbf{W}_m \mathbf{z}_m + \mathbf{b}_m)$. These projected features are then stacked into a sequence $\mathbf{H}^{(0)}_{\text{raw}} = [\tilde{\mathbf{z}}_img; \dots; \tilde{\mathbf{z}}_edge] \in \mathbb{R}^{4 \times d}$. To preserve the identity of each source, the sequence is augmented with a learnable modality positional encoding $\mathbf{P} \in \mathbb{R}^{4 \times d}$ to form the initialized input for the transformer blocks:
\begin{equation}
\mathbf{H}^{(0)} = \mathbf{H}^{(0)}_{\text{raw}} + \mathbf{P}
\end{equation}

\subsubsection{Cross-modal Interaction} 

We employ a two-layer Transformer encoder to model the complex dependencies between modalities. Each layer updates the sequence through multi-head self-attention and feed-forward networks:
\begin{equation}
\begin{aligned}
\mathbf{H}^{(l)} &= \mathrm{LN}(\mathbf{H}^{(l-1)} + \mathrm{MSA}(\mathbf{H}^{(l-1)})),  \\
\mathbf{H}^{(l)} &= \mathrm{LN}(\mathbf{H}^{(l)} + \mathrm{FFN}(\mathbf{H}^{(l)})),
\end{aligned}
\end{equation}
where $l \in \{1, 2\}$. This self-attention mechanism allows each modality token to adaptively aggregate semantic cues from others, facilitating a comprehensive cross-modal alignment.

\subsubsection{Aggregation and Robust Training}

The refined sequence $\mathbf{H}^{(2)}$ is compressed into a single vector through global average pooling across the modality dimension: $\mathbf{h}_{\text{agg}} = \frac{1}{4} \sum_{i=1}^{4} \mathbf{H}^{(2)}_i$. A residual MLP then projects this aggregate into the final fused embedding $\mathbf{z}_{\text{fus}}$.

To enhance the model's robustness against missing data, we implement a Modality Masking strategy during training, where one modality token in each batch is randomly zeroed out. This forces the encoder to maintain high-fidelity fusion even under incomplete observations. The entire module is optimized using the Similarity-Category Masked (SCM) loss, ensuring intra-class compactness in the fused semantic space.

\subsection{Shared-Trunk \& Heads Alignment (STH-Align)}
A prevalent approach to aligning EEG embeddings with image embeddings is to employ a diffusion prior. However, diffusion-based models typically require large-scale datasets; when only tens of thousands of EEG–image pairs are available, they tend to over-fit and incur heavy inference costs. We therefore propose Shared-Trunk \& Heads Alignment \textbf{(STH-Align)}, a lightweight module that simultaneously refines embeddings from four modalities by integrating multi-modal information to produce a single, refined output.

\subsubsection{Architecture}  



STH-Align consists of two primary components: a Shared-Trunk and Modality-Specific Heads. 

~\textbf{Shared-Trunk.}  
For a mini-batch we concatenate the four EEG embeddings  
$\mathbf x_{\text{cat}}=[\mathbf e^{\,\text{img}};\,\mathbf e^{\,\text{txt}};\,\mathbf e^{\,\text{depth}};\,\mathbf e^{\,\text{edge}}]\in\mathbb R^{4d}$  
and feed it into a 4-block MLP  
\begin{equation}
\mathbf h_{\ell}=\text{SiLU}\bigl(\text{LayerNorm}(\mathbf W_{\ell}\mathbf h_{\ell-1}+\mathbf b_{\ell})\bigr),\quad \ell=1,\dots,4,
\end{equation} 
yielding a common representation $\mathbf f\in\mathbb R^{d}$.  
This forces different modalities to share high-level statistics while suppressing modality-specific noise.

~\textbf{Modality-Specific Heads.}  
Each modality owns a 2-layer MLP projection head
\begin{equation}
\hat{\mathbf e}^{\,m}=\text{Proj}^{m}(\mathbf f),\quad m\in\{\text{img},\text{txt},\text{depth},\text{edge}\},
\end{equation}
whose output is L2-normalised so that subsequent cosine similarities are numerically stable.

\subsubsection{Objective}  
Given target image embeddings $\mathbf v^{m}$ produced by a frozen image encoder, STH-Align minimizes the total loss $\mathcal L_{\text{STH}}$, defined as:
\begin{equation}
\begin{aligned}
\mathcal L_{\text{STH}} = \sum_{m} \Bigl[ & \lambda_{\text{mse}} \underbrace{\|\hat{\mathbf e}^{\,m} - \mathbf v^{m}\|_2^{2}}_{\text{MSE Loss}} \\
& + \lambda_{\text{cos}} \underbrace{(1 - \cos(\hat{\mathbf e}^{\,m}, \mathbf v^{m}))}_{\text{Cosine Loss}} \\
& + \lambda_{\text{reg}} \underbrace{\|\hat{\mathbf e}^{\,m}\|_2^{2}}_{\text{Regularization}} \Bigr].
\end{aligned}
\end{equation}
The three terms balance Euclidean fidelity, angular alignment, and representation stability.

\subsubsection{Training \& Inference}  
STH-Align is trained separately with the EEG and Fusion encoders. modality dropout is applied so that one modality is randomly discarded per batch, encouraging the shared trunk to produce robust representations. At test time only the query modality's EEG embedding is processed; the refined embedding is used for retrieval or fed to the generative model. The shared-private design yields superior alignment quality while remaining lightweight, providing a compact and consistent cross-modal representation for the subsequent generation stage.

\subsection{SDXL Turbo + IP Adapter Generation}
Using the multimodal representations aligned by STH-Align, we adopt a parallel injection approach with SDXL-Turbo\cite{sauer2024adversarial} as the backbone diffusion model. To minimize potential uncertainty, the text modality is excluded, and we instead utilize three IP-Adapter\cite{ye2023ip} branches to simultaneously integrate EEG-derived image, depth, and edge cues.

\paragraph{Condition Branches}
Each modality is processed by an independent IP-Adapter-Layout variant to ensure consistent spatial-layout modeling. Specifically, the \textbf{image branch} captures global semantics and the overall layout, while the \textbf{depth branch} reinforces 3-D structural consistency, and the \textbf{edge branch} provides precise contour and shape details. These features are inserted in parallel into the Cross-Attention layers of the U-Net and aggregated using modality-specific weights.

\begin{table*}[t]
\centering
\small
\setlength{\tabcolsep}{1pt} 
\caption{Zero-shot retrieval performance on the Things-EEG dataset. The table reports the Top-1 and Top-5 accuracies (\%) for ten subjects and their average. \textbf{CogCapPro (I), (T), (D), (E), and (F)} denote the results using Image, Text, Depth, Edge, and multi-modal Fusion, respectively. Bold values indicate the best performance.}
\begin{tabular}{l@{\hspace{1pt}}ccccccccccccccccccccl@{\hspace{1pt}}l}
\toprule
\multirow{2}{*}{Method} & \multicolumn{2}{c}{Subject 1} & \multicolumn{2}{c}{Subject 2} & \multicolumn{2}{c}{Subject 3} & \multicolumn{2}{c}{Subject 4} & \multicolumn{2}{c}{Subject 5} & \multicolumn{2}{c}{Subject 6} & \multicolumn{2}{c}{Subject 7} & \multicolumn{2}{c}{Subject 8} & \multicolumn{2}{c}{Subject 9} & \multicolumn{2}{c}{Subject 10} & \multicolumn{2}{c}{Avg} \\ 
\cmidrule(lr){2-3} \cmidrule(lr){4-5} \cmidrule(lr){6-7} \cmidrule(lr){8-9} \cmidrule(lr){10-11} \cmidrule(lr){12-13} \cmidrule(lr){14-15} \cmidrule(lr){16-17} \cmidrule(lr){18-19} \cmidrule(lr){20-21} \cmidrule(lr){22-23}
                        & top-1 & top-5 & top-1 & top-5 & top-1 & top-5 & top-1 & top-5 & top-1 & top-5 & top-1 & top-5 & top-1 & top-5 & top-1 & top-5 & top-1 & top-5 & top-1 & top-5 & top-1 & top-5 \\ 
\midrule
BraVL\cite{du2023decoding}                   & 6.1   & 17.9  & 4.9   & 14.9  & 5.6   & 17.4  & 5.0   & 15.1  & 4.0   & 13.4  & 6.0   & 18.2  & 6.5   & 20.4  & 8.8   & 23.7  & 4.3   & 14.0  & 7.0   & 19.7  & 5.8   & 17.5  \\
NICE\cite{song2024decoding}                    & 13.2  & 39.5  & 12.5  & 40.3  & 14.5  & 42.7  & 15.3  & 39.6  & 10.1  & 31.5  & 16.5  & 44.0  & 17.0  & 42.1  & 12.9  & 56.1  & 15.4  & 41.6  & 17.4  & 45.8  & 14.1  & 43.6  \\
NICE-S\cite{song2024decoding}                  & 13.3  & 40.2  & 13.1  & 36.1  & 15.3  & 42.6  & 20.9  & 52.0  & 9.8   & 34.4  & 14.2  & 42.4  & 17.9  & 43.6  & 28.2  & 50.2  & 14.4  & 38.7  & 16.0  & 42.8  & 16.7  & 41.7  \\
NICE-G\cite{song2024decoding}                  & 15.2  & 40.1  & 13.9  & 40.1  & 15.7  & 42.7  & 17.6  & 48.9  & 9.0   & 29.7  & 16.3  & 44.4  & 14.9  & 43.1  & 20.3  & 52.1  & 14.1  & 39.7  & 19.6  & 46.7  & 15.6  & 42.8  \\
MB2C\cite{wei2024mb2c}                    & 23.6  & 56.3  & 22.6  & 50.5  & 26.3  & 60.1  & 34.8  & 67.0  & 21.3  & 53.0  & 31.0  & 62.3  & 35.0  & 64.8  & 39.0  & 69.3  & 27.5  & 59.3  & 33.1  & 70.8  & 28.4  & 60.3  \\
ATM-S\cite{li2024visual}                   & 23.6  & 50.4  & 22.0  & 54.5  & 24.0  & 62.4  & 31.4  & 60.9  & 12.9  & 43.0  & 21.4  & 51.1  & 20.5  & 51.5  & 38.8  & 72.0  & 34.4  & 51.5  & 29.1  & 63.5  & 25.5  & 60.4  \\
CogCap\cite{zhang2025cognitioncapturer}                  & 31.4  & 79.6  & 31.4  & 77.8  & 38.1  & 85.6  & 40.3  & 85.8  & 24.4  & 66.3  & 34.8  & 78.7  & 34.6  & 80.9  & 48.1  & 88.6  & 37.4  & 79.3  & 35.5  & 79.2  & 35.6  & 80.2  \\
VE-SDN\cite{chen2024visual}                  & 32.6  & 63.7  & 34.4  & 69.9  & 38.7  & 73.5  & 39.8  & 72.0  & 29.4  & 58.6  & 34.5  & 68.8  & 34.5  & 68.3  & 49.3  & 79.8  & 39.0  & 69.6  & 39.8  & 75.3  & 37.2  & 69.9  \\
UBP\cite{wu2025bridging}                     & 41.2  & 70.5  & 51.2  & 80.9  & 51.2  & 82.0  & 51.1  & 76.9  & 42.2  & 72.8  & 57.5  & 83.5  & 49.0  & 79.9  & 58.6  & 85.8  & 45.1  & 76.2  & 61.5  & 88.2  & 50.9  & 79.7  \\
ATS\cite{wu2025shrinking}                     & 53.0  & 79.0  & \textbf{62.0} & 87.5  & 61.5  & 89.0  & 57.0  & 86.5  & \textbf{55.0} & \textbf{84.0} & \textbf{68.0} & 90.5  & 53.0  & 84.0  & 66.5  & 91.0  & \textbf{58.5} & 86.0  & \textbf{67.5} & 89.0  & 60.2  & 86.7  \\ 
\midrule
CogCapPro(I)            & 54.2  & 83.2  & 50.5  & 82.9  & 49.5  & 85.6  & 56.5  & 85.4  & 40.9  & 74.0  & 52.3  & 86.2  & 50.8  & 79.5  & 63.1  & 90.0  & 49.5  & 81.7  & 59.3  & 86.5  & 52.7  & 83.5  \\
CogCapPro(T)            & 10.9  & 31.9  & 13.0  & 34.6  & 12.3  & 36.2  & 17.4  & 43.6  & 11.4  & 30.5  & 15.0  & 40.3  & 15.4  & 40.8  & 17.3  & 44.9  & 13.3  & 38.0  & 16.0  & 44.8  & 14.2  & 38.6  \\
CogCapPro(D)            & 17.8  & 47.0  & 15.0  & 40.8  & 19.7  & 47.3  & 19.7  & 51.2  & 12.1  & 34.8  & 15.8  & 42.2  & 18.8  & 44.5  & 20.2  & 47.9  & 17.5  & 43.4  & 18.5  & 43.9  & 17.5  & 44.3  \\
CogCapPro(E)            & 29.2  & 62.9  & 27.3  & 64.0  & 32.8  & 65.6  & 36.8  & 68.8  & 24.3  & 56.1  & 34.8  & 68.1  & 25.5  & 61.1  & 35.3  & 71.7  & 24.1  & 59.1  & 29.1  & 66.4  & 29.9  & 64.4  \\
CogCapPro(F)            & \textbf{61.9} & \textbf{91.5} & 59.4  & \textbf{90.9} & \textbf{62.2} & \textbf{91.9} & \textbf{67.0} & \textbf{94.6} & 45.2  & 78.8  & 64.3  & \textbf{91.8} & \textbf{58.0} & \textbf{90.4} & \textbf{73.2} & \textbf{95.7} & 55.0  & \textbf{88.6} & 65.8  & \textbf{93.6} & \textbf{61.2} & \textbf{90.8} \\ 
\bottomrule
\end{tabular}
\label{tab:retrieval_results}
\end{table*}

\begin{table}[t]
\centering
\small
\setlength{\tabcolsep}{0.7pt} 
\caption{Zero-shot retrieval performance on the Things-MEG dataset.}
\resizebox{\columnwidth}{!}{ 
\begin{tabular}{l@{\hspace{0.7pt}}cccccccccc@{\hspace{0.7pt}}}
\toprule
\multirow{2}{*}{Method} &
\multicolumn{2}{c}{Subject 1} &
\multicolumn{2}{c}{Subject 2} &
\multicolumn{2}{c}{Subject 3} &
\multicolumn{2}{c}{Subject 4} &
\multicolumn{2}{c}{Avg} \\
\cmidrule(lr){2-3}\cmidrule(lr){4-5}\cmidrule(lr){6-7}\cmidrule(lr){8-9}\cmidrule(lr){10-11}
& top-1 & top-5 & top-1 & top-5 & top-1 & top-5 & top-1 & top-5 & top-1 & top-5 \\
\midrule
NICE\cite{song2024decoding}                    & 9.6  & 27.8 & 18.5 & 47.8 & 14.2 & 41.6 & 9.0  & 26.6 & 12.8 & 36.0 \\
NICE-S\cite{song2024decoding}                  & 9.8  & 27.8 & 18.6 & 46.4 & 10.5 & 38.4 & 11.7 & 27.2 & 12.7 & 35.0 \\
NICE-G\cite{song2024decoding}                  & 8.7  & 30.5 & 21.8 & 56.6 & 16.5 & 49.7 & 10.3 & 32.3 & 14.3 & 42.3 \\
UBP\cite{wu2025bridging}                     & 15.0 & 38.0 & 46.0 & 80.5 & 27.3 & 59.0 & 18.5 & 43.5 & 26.7 & 55.2 \\
ATS\cite{wu2025shrinking}                     & 15.5 & 42.0 & \textbf{61.5} & 88.0 & 33.0 & 69.5 & 19.0 & 43.5 & \textbf{32.3} & 62.4 \\
\midrule
CogCapPro(I)            & \textbf{17.2} & 43.0 & 49.5 & 83.0 & 34.5 & 68.5 & 20.0 & 47.0 & 30.3 & 60.4 \\
CogCapPro(T)            & 7.5  & 23.7 & 10.5 & 36.5 & 12.3 & 31.2 & 8.5  & 21.7 & 9.7  & 28.3 \\
CogCapPro(D)            & 7.0  & 19.7 & 13.0 & 35.5 & 12.5 & 31.8 & 6.2  & 18.2 & 9.7  & 26.3 \\
CogCapPro(E)            & 9.0  & 25.5 & 27.0 & 54.0 & 16.3 & 45.7 & 11.0 & 29.5 & 15.8 & 38.7 \\
CogCapPro(F)            & 15.0 & \textbf{45.0} & 54.5 & \textbf{89.5} & \textbf{36.3} & \textbf{72.2} & \textbf{21.5} & \textbf{51.7} & 31.8 & \textbf{64.6} \\
\bottomrule
\end{tabular}
}
\label{tab:retrieval_results_meg}
\end{table}

\paragraph{Joint Denoising and Reconstruction}
During SDXL-Turbo denoising, key-value pairs from all adapters undergo simultaneous attention-based fusion. This multi-view integration directly guides noise prediction, ensuring the output is semantically and structurally consistent with the EEG-derived spatial and visual stimulus.

\section{Experiments} \label{sec:experiment}
\subsection{Datasets and Preprocessing}

\subsubsection{Datasets}
We evaluated CogCapPro on two large-scale brain signal datasets: THINGS-EEG~\cite{gifford2022large} and THINGS-MEG~\cite{hebart2023things}. The THINGS-EEG dataset contains EEG recordings from 10 subjects under a rapid serial visual presentation paradigm\cite{keysers2001speed}. The training set comprises 1,654 conceptual categories with 10 images per category, each image repeated 4 times per subject. The test set includes 200 conceptual categories with 1 image per category, each image repeated 80 times per subject.The THINGS-MEG dataset involves MEG recordings from 4 participants with 271 channels. The training set consists of 1,854 conceptual categories $\times$ 12 images $\times$ 1 repetition, while the test set contains 200 conceptual categories $\times$ 1 image $\times$ 12 repetitions. All experimental procedures were approved by the relevant institutional review boards, and informed consent was obtained from all participants prior to their involvement in the study.

\subsubsection{Data Preprocessing and Multi-modal Expansion}

For both EEG and MEG signals, we follow the preprocessing process described in~\cite{wu2025bridging} to ensure experimental fairness. To provide richer supervision signals, we expand the original visual stimuli into four modalities. All images are resized to $224\times224$ pixels. We convert the original images into edge maps with salient contour features by applying the Canny edge detection algorithm to Gaussian-smoothed grayscale images. Additionally, depth maps are estimated using the DepthAnything model, and textual descriptions are generated via BLIP-2. This multi-modal expansion enables the model to address modality and fidelity misalignment by leveraging complementary information from the visual stimuli.

\subsubsection{Implementation Details}
We employ two OpenCLIP~\cite{ilharco_gabriel_2021_5143773} variants, ResNet-50 and ViT-H-14, as the base visual encoders. To evaluate performance, we compare them against four representative EEG encoders: Shallownet, Deepnet~\cite{schirrmeister2017deep}, EEGNet~\cite{song2022eeg}, and TSConv~\cite{song2024decoding}. 

All experiments were conducted on 8 NVIDIA RTX 3090 GPUs. We trained the models for 80 epochs using a batch size of 1024. To prevent overfitting, text modality training is capped at 30 epochs. Each experiment repeated 5 times to ensure statistical significance. For the retrieval alignment task, we used the AdamW optimizer with a learning rate of $1\times10^{-4}$. Each modal expert encoder is assigned an independent optimizer to prevent inter-modal information leakage. Regarding the STH-Align objective, the loss weights are set as $\lambda_{\text{mse}} = 1.0$, $\lambda_{\text{cos}} = 0.5$, and $\lambda_{\text{reg}} = 1\times10^{-4}$.

For the UM module, the memory bank update coefficient $\gamma$ and blur radius adjustment $c$ are set to 0.3 and 6, respectively. The top-$k$ parameter in the SCM-Loss is 10. To ensure comparability and reproducibility in EEG-to-image reconstruction, we follow the evaluation protocols established in~\cite{li2024visual, benchetritbrain}.

\begin{figure*}[t]
    \centering
    \includegraphics[width=1\linewidth]{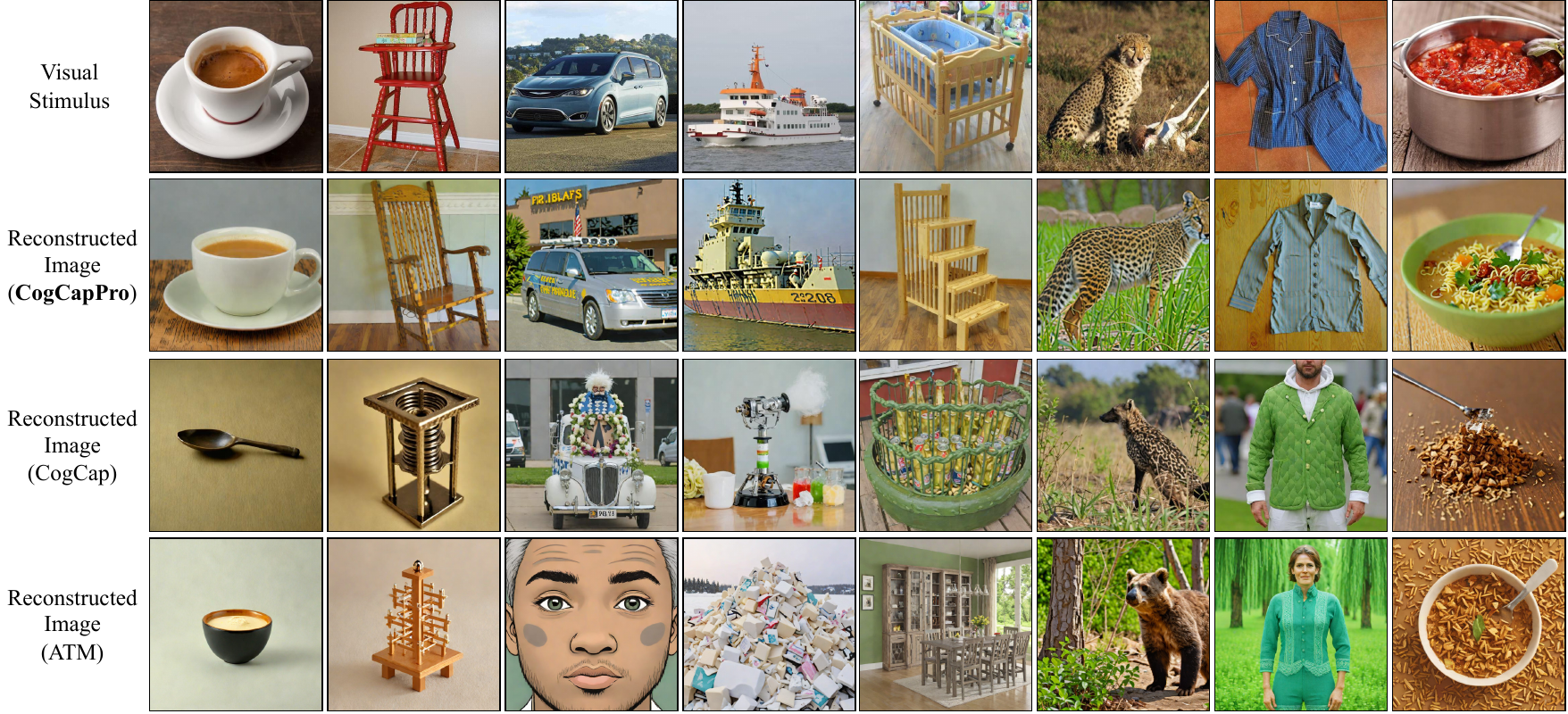}
    \caption{Qualitative comparison of image reconstruction results on the Things-EEG dataset. The first row displays the ground-truth visual stimuli. Subsequent rows show reconstructions generated by CogCapPro, CogCap, and ATM, respectively.}
    \label{fig:recon_results}
\vspace{-1em}
\end{figure*}


\begin{table}[t]
\centering
\small 
\setlength{\tabcolsep}{1pt} 
\caption{Image reconstruction performance. Metrics are divided into low-level and high-level features, where arrows indicate the direction of better performance.}
\begin{tabular}{lccccc}
\toprule
\multirow{2}{*}{Method} & \multicolumn{2}{c}{Low-level} & \multicolumn{3}{c}{High-level} \\ 
\cmidrule(r){2-3} \cmidrule(l){4-6} 
& PixCorr$\uparrow$ & SSIM$\uparrow$ & Inception$\uparrow$ & CLIP$\uparrow$ & SwAV$\downarrow$ \\ 
\midrule
MindEye (fMRI)\cite{scotti2024mindeye2} & 0.322 & 0.431 & 0.954 & 0.930 & 0.344 \\ \midrule
META (MEG)\cite{benchetritbrain} & 0.090 & 0.341 & 0.703 & 0.811 & 0.567 \\
ATM (MEG)\cite{li2024visual} & 0.104 & 0.340 & 0.619 & 0.603 & 0.651 \\ \midrule
CogCap (EEG)\cite{zhang2025cognitioncapturer} & 0.150 & 0.347 & 0.669 & 0.715 & 0.590 \\
ATM (EEG)\cite{li2024visual} & 0.160 & 0.345 & 0.734 & 0.786 & 0.582 \\
\textbf{CogCapPro} (EEG) & \textbf{0.163} & \textbf{0.398} & \textbf{0.779} & \textbf{0.830} & \textbf{0.553} \\ 
\bottomrule
\end{tabular}
\label{tab:reconstruction_results}
\end{table}

\subsection{Retrieval Performance}

This section evaluates the retrieval performance of the CognitionCapturerPro \textbf{(hereafter abbreviated as CogCapPro)} model. We conducted extensive experiments on both the Things-EEG and Things-MEG datasets, comparing our approach with state-of-the-art methods. The comprehensive results are summarized in Table~\ref{tab:retrieval_results} and Table~\ref{tab:retrieval_results_meg}.

Regarding the within-subject experiments on the Things-EEG dataset, CogCapPro achieves significant performance improvements over existing leading methods. Specifically, it reaches a Top-1 accuracy of 61.2\% and a Top-5 accuracy of 90.8\%, surpassing the comparative method ATS~\cite{wu2025shrinking} by 1.0\% and 4.1\%, respectively. 

The results on the Things-MEG dataset follow a similar trend. In the 200-class zero-shot retrieval task—where the random guess accuracy is only 0.5\%—CogCapPro yields highly competitive outcomes. Its Top-1 accuracy reaches 31.8\% and its Top-5 accuracy is 64.6\%, maintaining a 2.2\% lead over ATS~\cite{wu2025shrinking} in the more representative Top-5 metric. 

To further investigate the contribution of individual modalities, Table~\ref{tab:retrieval_results} and Table~\ref{tab:retrieval_results_meg} provide a detailed breakdown of CogCapPro's performance across different inputs. The data confirms that the multi-modal fusion strategy achieves the highest retrieval accuracy, validating the effectiveness of our joint representation approach. Among the single modalities, the image modality contributes most significantly, with performance second only to the fusion result, followed by the edge, depth, and text modalities. This ranking clearly illustrates the relative importance of each modality in the cross-modal retrieval task.

\subsection{Reconstruction Performance}

On the Things-EEG dataset, we evaluated image reconstruction using five metrics—PixCorr, SSIM, Inception, CLIP, and SwAV—to comprehensively assess the reconstruction quality from both low-level visual features and high-level semantic features. As shown in Table \ref{tab:reconstruction_results}, CogCapPro significantly outperforms existing methods in reconstructing both low-level and high-level features. Specifically, at the semantic level, CogCapPro achieves a 4.4\% improvement on the CLIP metric and reduces the SwAV metric (where a lower value indicates better performance) by 2.9\%, demonstrating its superior capability in capturing complex neural-to-visual mappings and achieving highly competitive performance in EEG-based image reconstruction.

Figure \ref{fig:recon_results} visualizes representative reconstruction examples. For the ``cheetah" stimulus, ATM \cite{li2024visual} only captures the coarse ``animal" category (reconstructing a bear), while CogCap \cite{zhang2025cognitioncapturer} improves granularity to a ``spotted hyena." In contrast, CogCapPro accurately recovers the ``cheetah" and its distinctive spot patterns, demonstrating superior fine-grained semantic alignment. Similarly, for the ``coffee cup," ATM fails to resolve the ``handle" and ``saucer" structures, and CogCap yields an incorrect ``spoon." CogCapPro, however, successfully restores the complete object geometry, highlighting its precision in mapping neural signals to complex structural details.

\begin{table}[t]
\centering
\small
\setlength{\tabcolsep}{4pt}
\caption{Ablation study of different modules on the baseline model.}
\begin{tabular}{ccccccc}
\toprule
\multirow{2}{*}{Baseline} & \multirow{2}{*}{UM} & \multirow{2}{*}{Loss} & \multirow{2}{*}{Mask} & \multicolumn{2}{c}{Avg (Fusion)} \\
\cmidrule(lr){5-6}
& & & & top-1 (std) & top-5 (std) \\
\midrule
\ding{51} & \ding{55} & \ding{55} & \ding{55} & 51.8 (5.7) & 84.8 (4.8) \\ 
\ding{51} & \ding{51} & \ding{55} & \ding{55} & 54.7 (6.4) & 87.1 (3.9) \\
\ding{51} & \ding{51} & \ding{51} & \ding{55} & 60.7 (6.2) & 90.4 (3.2) \\
\ding{51} & \ding{51} & \ding{51} & \ding{51} & 61.2 (7.2) & 90.8 (4.4) \\
\bottomrule
\end{tabular}
\label{tab:baseline_config}
\end{table}


\begin{table}[t]
\centering
\small
\setlength{\tabcolsep}{2pt} 
\caption{Performance comparison across different EEG and image encoders.}
\begin{tabular}{lccc}
\toprule
\multirow{2}{*}{EEG Encoder} & \multirow{2}{*}{Image Encoder} & \multicolumn{2}{c}{Avg (Fusion)} \\
\cmidrule(lr){3-4} 
                  &               & top-1 (std)    & top-5 (std)    \\ 
\midrule
CogCap Encoder    & Vit-H-14      & 56.0 (6.7)     & 87.6 (5.3)     \\
CogCap Encoder    & RN50          & 61.2 (7.2)     & 90.8 (4.4)     \\
ShallowNet        & RN50          & 37.0 (7.3)     & 72.6 (8.2)     \\
DeepNet           & RN50          & 24.4 (4.6)     & 56.8 (7.7)     \\
TSConv            & RN50          & 44.7 (7.5)     & 79.9 (6.5)     \\
EEGNet            & RN50          & 31.0 (6.6)     & 65.2 (8.6)     \\
\bottomrule
\end{tabular}
\label{tab:encoder_performance}
\end{table}

\subsection{Ablation Studies}

We evaluate CogCapPro's robustness through systematic ablations of its architectural modules, target brain regions, and EEG frequency bands. The specific details are as follows:

\subsubsection{Module Ablation Analysis}
We sequentially introduced the UM, the SCM-Loss, and the Modality Mask mechanism on top of the baseline model to progressively investigate the impact of each module on retrieval performance. As shown in Table \ref{tab:baseline_config}, both Top-1 and Top-5 accuracy showed continuous improvement with the incremental addition of each module, indicating that all three components effectively enhance model performance. Among them, the SCM-Loss module contributed the most to performance gain, increasing the Top-1 accuracy by 6.0\%, establishing it as a key optimization component in the architecture.

\subsubsection{Impact of Encoder Selection}
We further compared the impact of different EEG encoders and image encoders on model performance. The results are shown in Table \ref{tab:encoder_performance}.

\textbf{EEG Encoder Comparison:} Evaluation of EEG backbones (Table \ref{tab:encoder_performance}) reveals that CogCap Encoder leads the performance, yielding 61.2\% Top-1 and 90.8\% Top-5 accuracy. This marks a notable improvement over TSConv (44.7\% Top-1) and more than doubles the performance of DeepNet (24.4\%), validating our encoder’s efficiency in extracting high-level semantic information from complex EEG signals.

\textbf{Image Encoder Comparison:} With the CogCap Encoder as the fixed EEG backbone, RN50 significantly outperformed ViT-H-14 as the image encoder, achieving Top-1/Top-5 accuracies of 61.2\%/90.8\% compared to 56.0\%/87.6\%. We posit that this performance gap stems from the relative sparsity of visual information within EEG signals; specifically, the feature distribution of RN50 may align more effectively with the information density of EEG representations than ViT-H-14.

\begin{figure*}
    \centering
    \includegraphics[width=1\linewidth]{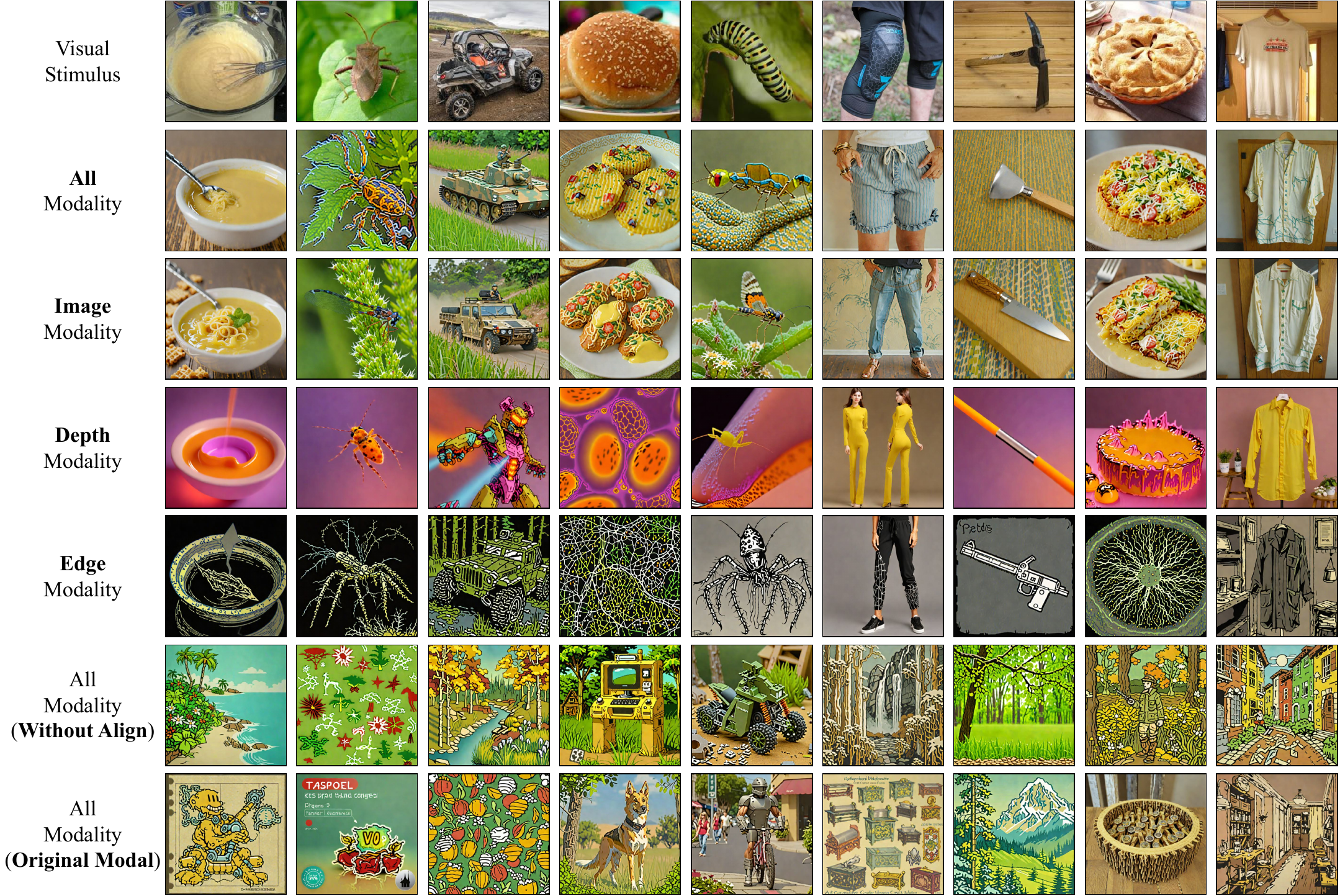}
    \caption{Qualitative comparison of reconstruction results across different modality configurations and alignment strategies. From top to bottom: the original visual stimulus, results from all modalities, single modalities, and variants without our alignment module. Our full-modality approach effectively integrates complementary features to produce the most consistent reconstructions.}
    \label{fig:comparemodality}
\end{figure*}


\begin{table}[t]
\centering
\small
\setlength{\tabcolsep}{1pt} 
\caption{Ablation study of modality combinations and alignment modules.}
\begin{tabular}{lccccc}
\toprule
\multirow{2}{*}{Method} & \multicolumn{2}{c}{Low-level}      & \multicolumn{3}{c}{High-level}                          \\
\cmidrule(lr){2-3} \cmidrule(lr){4-6} 
                  & PixCorr$\uparrow$ & SSIM$\uparrow$ & Inception$\uparrow$ & CLIP$\uparrow$ & SwAV$\downarrow$ \\
\midrule
Baseline Align   & 0.097             & 0.192          & 0.481               & 0.505          & 0.760            \\
w/o Align Module & 0.103             & 0.215          & 0.516               & 0.483          & 0.751            \\
\midrule
CogCapPro (Image)  & 0.141             & 0.317          & 0.773               & \textbf{0.833} & 0.558            \\
CogCapPro (Depth)  & 0.122             & \textbf{0.407} & 0.636               & 0.643          & 0.663            \\
CogCapPro (Edge)   & 0.029             & 0.130          & 0.645               & 0.658          & 0.735            \\
CogCapPro (All)    & \textbf{0.163}    & 0.398          & \textbf{0.779}      & 0.830          & \textbf{0.553}   \\
\bottomrule
\end{tabular}
\label{tab:cogcappro_variants}
\end{table}

\begin{figure*}
    \centering
    \includegraphics[width=1\linewidth]{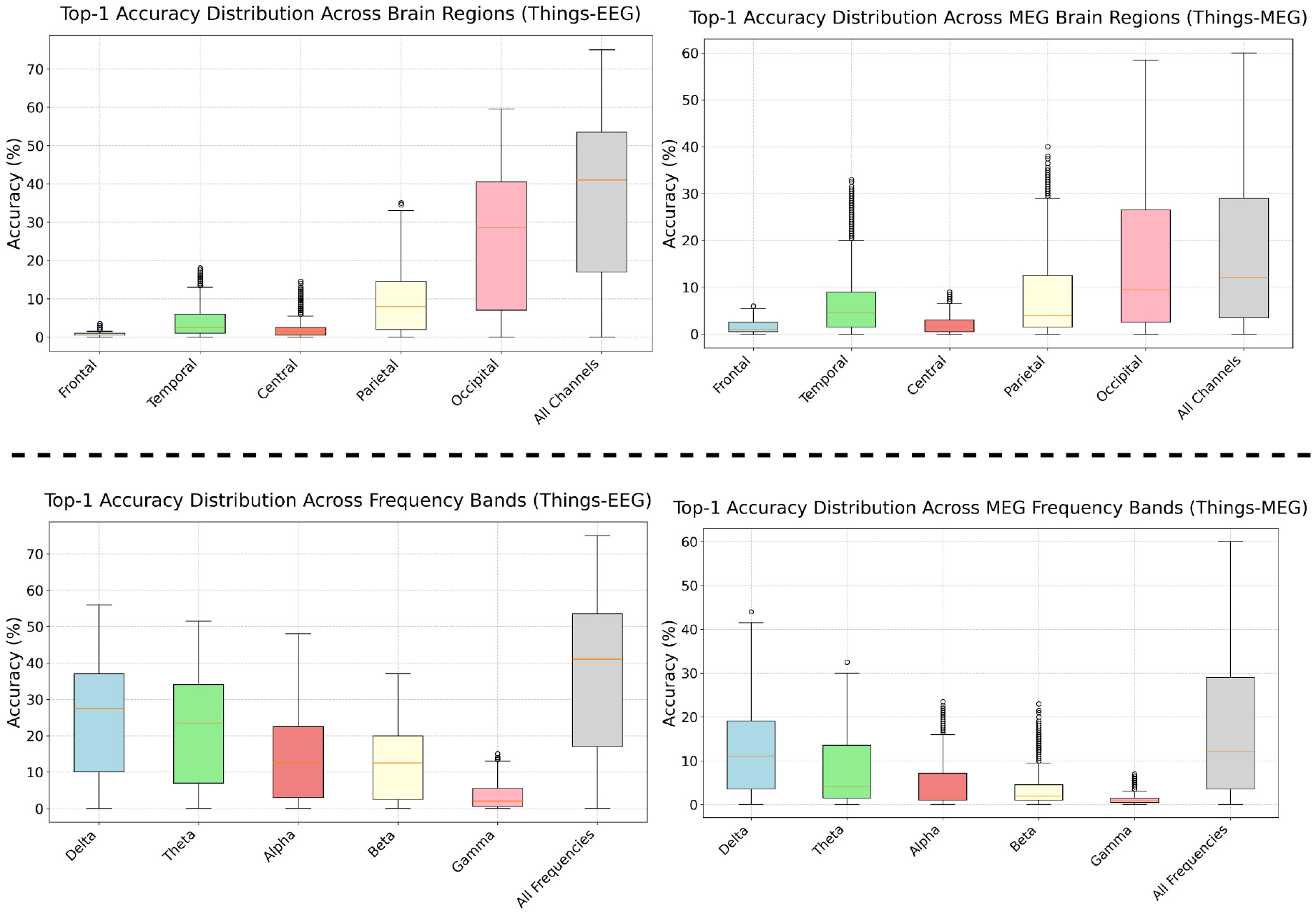}
    \caption{Top-1 accuracy distribution across spatial and spectral dimensions. The top row illustrates the classification performance across different brain regions (Frontal, Temporal, Central, Parietal, and Occipital), while the bottom row shows the performance across five frequency bands (Delta, Theta, Alpha, Beta, and Gamma). Left and right columns represent results for the Things-EEG and Things-MEG datasets, respectively. ``All Channels" denotes the integration of all spatial or spectral information.}
    \label{fig:eeg_meg_boxplot}
\end{figure*}

\subsubsection{Impact of Different Modalities on Reconstruction}

To isolate the contribution of each modality, we independently evaluated the image, depth, and edge features, while specifically sidelining the text modality due to its limited discriminative power observed in prior retrieval phases. As evidenced by the quantitative data in Table \ref{tab:cogcappro_variants} and Figure \ref{fig:comparemodality}, while specific single-modality models exhibit performance comparable to the full-modality set in isolated metrics like SSIM (with marginal fluctuations below 0.0X\%), the complete CogCapPro architecture maintains a distinct advantage in overall stability and cross-modal synergy, effectively mitigating the limitations inherent in any single information source.

Qualitatively, while the image modality captures holistic content, depth and edge modalities prioritize structural contours and textures at the expense of color fidelity. As illustrated in Figure \ref{fig:comparemodality}, the full-modality reconstruction yields more accurate scale for the ``bug" stimulus. Notably, in the ``bowl of butter" case, the image-only variant introduces semantic hallucinations (e.g., ``noodles"), which are effectively suppressed by the multi-modal fusion to restore a realistic structure. These findings demonstrate that our fusion strategy leverages cross-modal complementarity to enhance reconstruction stability and suppress noise.

\subsubsection{Role of the Alignment Module}
As shown in Table \ref{tab:cogcappro_variants} and Figure \ref{fig:comparemodality}, excluding the alignment module or employing only the baseline alignment method resulted in reconstructed images with noticeable semantic noise and structural inconsistencies. This led to a measurable performance degradation compared to our proposed method. These results highlight the effectiveness of the STH-Align module in facilitating precise feature semantic alignment.

\subsubsection{Impact of Brain Regions and Frequency Bands}
To explore the impact of brain regions and frequency bands on visual recognition tasks, we conducted a systematic analysis based on EEG and MEG data from both spatial (brain regions) and spectral (frequency bands) dimensions. The results are illustrated in Figure \ref{fig:eeg_meg_boxplot}, which presents four sets of box plots illustrating the Top-1 accuracy distributions (EEG/MEG $\times$ Spatial/Spectral).

\textbf{Spatial Dimension:} The Occipital lobe showed the highest recognition accuracy in both EEG and MEG, consistent with the neuroscientific conclusion that the ``occipital lobe serves as the primary visual cortex." In contrast, the Frontal lobe performed poorly in this task, as it is primarily involved in higher cognitive functions and is less associated with perceptual-level tasks. The advantage of EEG in the occipital region was more pronounced than that of MEG, potentially stemming from EEG's higher sensitivity to potentials on the cortical surface.

\textbf{Spectral Dimension:} Low-frequency bands (Delta: 0--4 Hz, Theta: 4--8 Hz) performed better in the recognition task than the high-frequency Gamma band (50--100 Hz), aligning with the neural mechanism that low-frequency activity participates in perceptual encoding. The Gamma band is susceptible to noise interference, resulting in lower information validity. Integrating all frequency bands (All Channels) achieved the optimal performance, indicating that the synergistic integration of multiple frequency bands helps enhance the model's recognition capability.

\textbf{Cross-Dimension Analysis:} The overall recognition accuracy of EEG across dimensions was higher than that of MEG, possibly because EEG is more suitable for capturing cortical surface activities related to visual tasks. Furthermore, weaker-performing brain regions or frequency bands (e.g., Frontal lobe, Gamma band) exhibited greater dispersion in accuracy distribution, reflecting the variability of individual neural activity; whereas the results for the Occipital lobe and All Channels were more concentrated, indicating the consistency of key neural circuits across the population.

\section{Interpretability Analysis}

To assess the interpretability of CogCapPro, this section investigates its internal mechanisms through semantic analysis, retrieval visualization, and Grad-CAM-based brain topography. These visualizations demonstrate the efficacy of our multi-modal fusion in semantic representation while validating the model's biological plausibility by reflecting a decoding process consistent with known cortical activity and visual attention.

\subsection{Semantic Analysis and Retrieval Visualization}

\begin{figure*}
    \centering
    \includegraphics[width=1\linewidth]{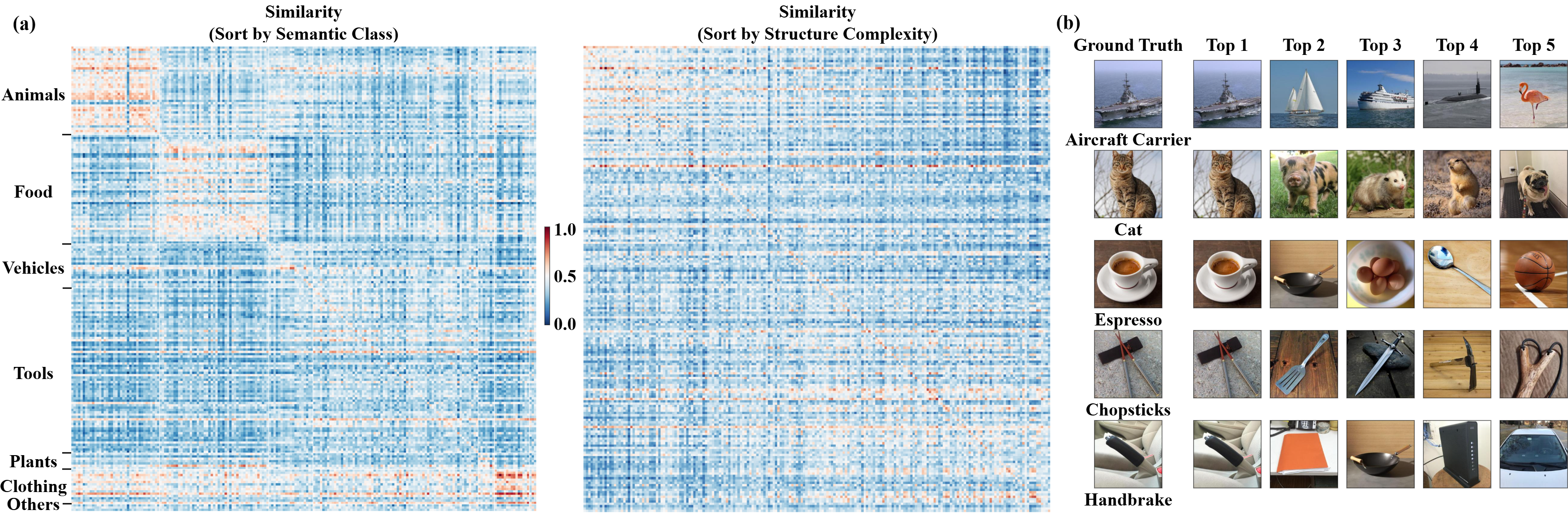}
    \caption{Semantic Analysis and Retrieval Visualization. (a) Similarity heatmaps of the fusion modality embeddings, sorted by semantic categories (left) and structural complexity from low to high (right). (b) Top-5 retrieval results for representative stimulus images.}
    \label{fig:Retrievalvis}
\end{figure*}

First, we conduct similarity analysis at the semantic level. Specifically, we divide the dataset into seven semantic categories (animals, food, vehicles, tools and equipment, plants, clothing, others) and compute similarity heatmaps for embeddings from different modalities. For simplicity, we refer to the fused embedding as the fusion modality. As illustrated in Figure \ref{fig:Retrievalvis}, the similarity heatmaps for depth and edge modalities exhibit indistinct patterns, indicating that these modalities are relatively insensitive to high-level semantic information. This is consistent with intuitive knowledge—depth and edges primarily capture structural topology rather than abstract semantics. In contrast, the image, text, and fusion modalities demonstrate clear discriminability for different semantic categories, with the fusion modality performing the best, thereby validating the effectiveness of our proposed fusion strategy in integrating multi-modal semantic information.

Additionally, we perform similarity analysis by arranging images in order of structural complexity from low to high. In the image and fusion modalities, low-complexity images show high similarity, while high-complexity images demonstrate high inter-sample variance, reflecting the model's sensitive capture of intricate visual features (see Figure \ref{fig:Retrievalvis}). The text modality's heatmap shows no discernible trends, while the depth and edge modalities exhibit global similarity consistency, which we attribute to the relatively uniform distribution of contour and edge information across most images.

In terms of retrieval result visualization, we display the Top-5 classification results randomly selected from one participant in Figure \ref{fig:Retrievalvis}. The predicted results are highly consistent with the ground-truth labels at the semantic level; for example, ``cat" is retrieved as various animal species, ``aircraft carrier" is retrieved as different types of maritime vessels, and ``chopsticks" is retrieved as ``elongated objects." This further confirms the model's effective recognition of image content from EEG signals, providing both strong biological plausibility and empirical support for the feasibility of EEG-based image recognition tasks.

\begin{figure}[t]
    \centering
    \includegraphics[width=1\linewidth]{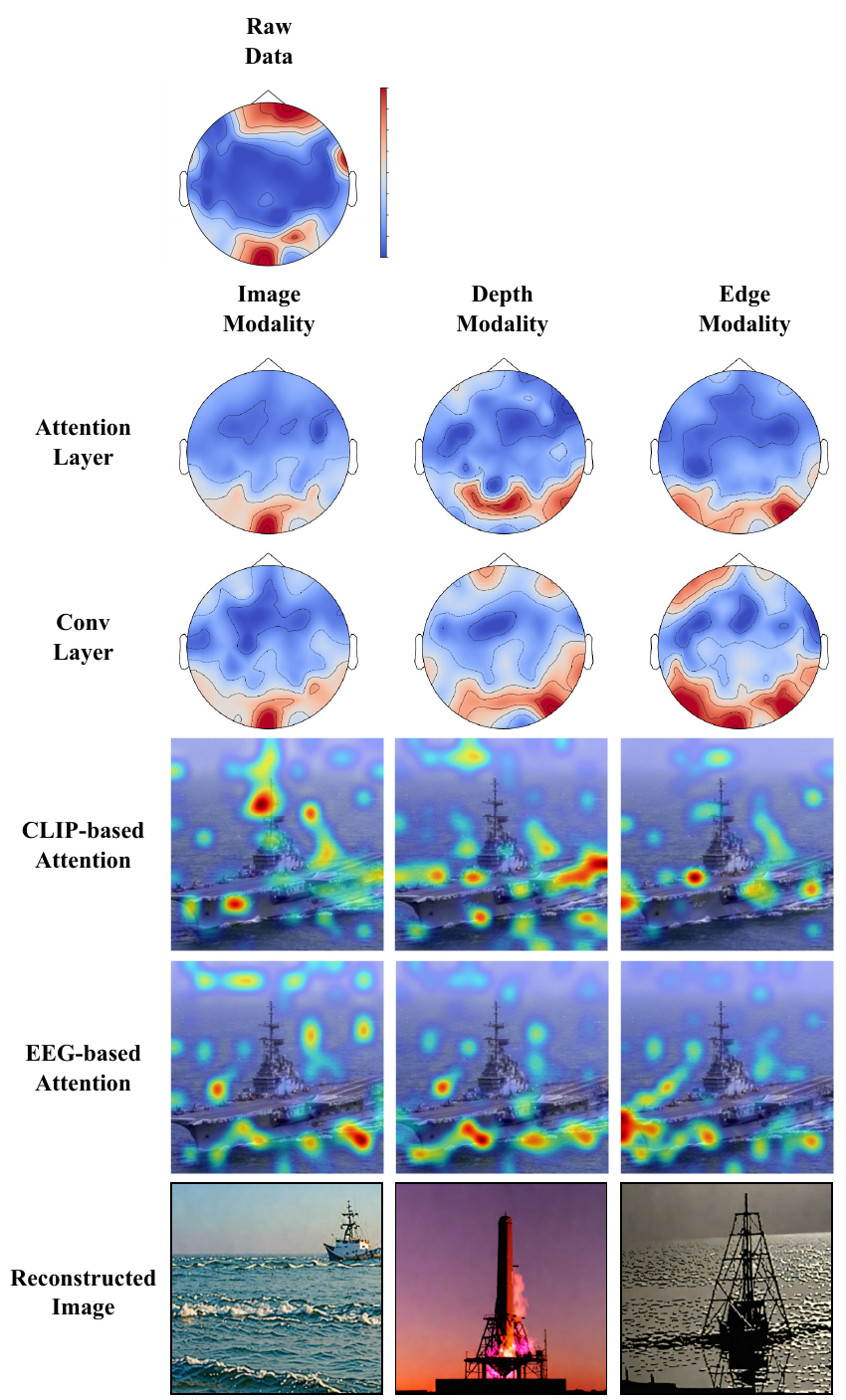}
    \caption{Visualization of model interpretability. (Top) Topographical maps of raw data and internal model layers (Attention/Convolution). (Middle) Grad-CAM heatmaps comparing CLIP and EEG-based attention across Image, Depth, and Edge modalities. (Bottom) Samples of reconstructed images.}
    \label{fig:topoattn}
\end{figure}

\subsection{Multi-Modal Interaction and Grad-CAM Exploration}

To enhance the model's interpretability, this section investigates the internal mechanisms of the CogCapPro framework by exploring the interactions among EEG signals, original stimuli, and reconstructed images. By employing Grad-CAM technology\cite{selvaraju2017grad} to generate brain topography maps and image heatmaps, we analyze the model’s attention mechanisms from both neural and visual perspectives, thereby validating its biological plausibility.

\subsubsection{Brain Topography Exploration} We first examine how the model identifies and utilizes brain regional activities relevant to image recognition. As illustrated in Figure~\ref{fig:topoattn}, the original EEG topography exhibits prominent activity in the frontal region, which is often associated with high-level cognitive noise or non-visual tasks\cite{fatourechi2007emg}. However, after processing through the convolution and attention modules, the model’s focus shifts significantly toward the occipital and temporal regions. These areas are recognized in neuroscience as core centers for visual object recognition, such as the V1 area for primary visual processing and the Inferior Temporal (IT) cortex for category-level representation~\cite{dicarlo2007untangling, dapello2023aligning}. The results demonstrate that CogCapPro effectively incorporates frontal lobe activity while selectively prioritizing specialized vision-related cortical information. The activation matches where visual processing occurs in the brain, confirming that the model relies on meaningful biological signals.

\subsubsection{Image Heatmap Analysis} Furthermore, we utilize Grad-CAM to visualize the attention areas of different modality encoders on the stimulus images, comparing them with the CLIP baseline. Using the ``warship" image as an example, Figure~\ref{fig:topoattn}, we observe distinct and complementary attention patterns across modalities: the image modality primarily focuses on salient semantic features (e.g., the bow), while the depth and edge modalities emphasize structural topology (e.g., the lower hull). This observation is consistent with the hierarchical processing of the human visual system, where edge detection occurs in early visual pathways (V1/V2) and comprehensive semantic integration involves higher layers (V4/IT)\cite{dicarlo2007untangling, hung2005fast}.

\subsubsection{Synthesis of Modalities} 


The ``warship" example illustrates how these modalities complement each other during reconstruction (Figure~\ref{fig:topoattn}). The image modality identifies key semantic features but struggles with spatial orientation, leading to distorted shapes. In contrast, depth and edge modalities maintain the correct structural layout and vessel orientation but provide no specific semantic information. CogCapPro resolves these individual limitations through multi-modal fusion, ensuring both structural and semantic consistency.

\begin{figure}
    \centering
    \includegraphics[width=1\linewidth]{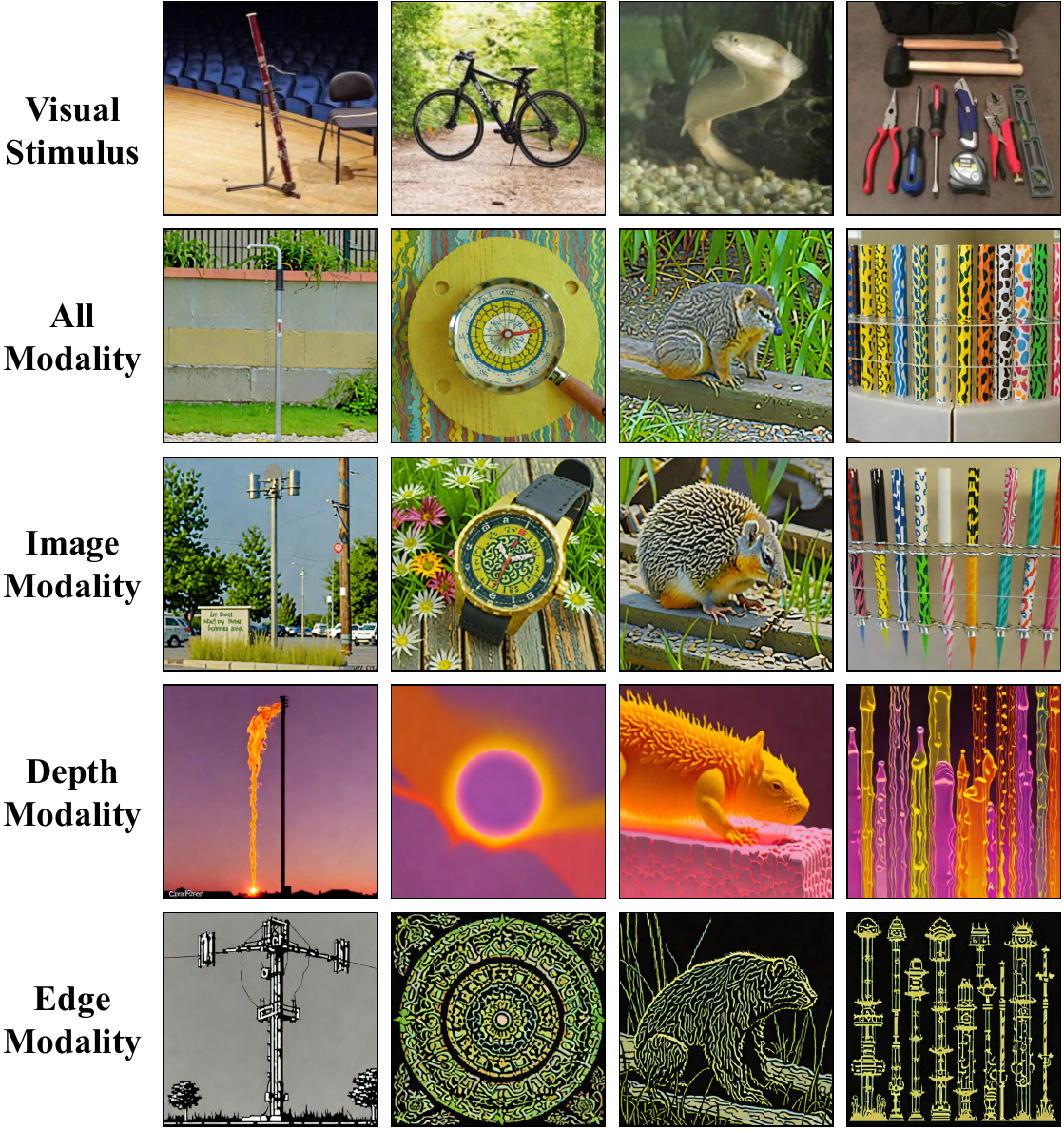}
    \caption{Visualization of reconstruction limitations. Rows from top to bottom display the original stimulus images and corresponding reconstructions generated using all modalities, image modality, depth modality, and edge modality, respectively.}
    \label{fig:failure}
\end{figure}

\section{Discussion and Conclusion}


CogCapPro addresses fidelity loss and representation shift in neural decoding by employing specialized alignment modules. These modules facilitate the integration of multimodal data, effectively synchronizing internal neural signals with external visual stimuli. Evaluations on the THINGS-EEG \cite{gifford2022large} and THINGS-MEG \cite{hebart2023things} datasets show that CogCapPro significantly outperforms the original CognitionCapturer \cite{zhang2025cognitioncapturer} and baseline methods in retrieval accuracy. By capturing both global semantics and local geometric features, this approach provides a refined technical foundation for BCI applications in medical rehabilitation and assistive technologies.

However, achieving real-time and efficient visual decoding still faces challenges. The first is the issue of response stability. Currently, models must follow experimental paradigms by averaging across multiple repeated stimuli to improve the signal-to-noise ratio, which limits the application scenarios of real-time visual neural decoding. Therefore, how to extract efficient features under single-trial conditions is a key research direction for the future.

Secondly, the depth of utilization of brain regions needs to be strengthened. Spatial analysis shows that decoding contributions primarily come from the occipital lobe as the primary visual cortex \cite{cichy2014resolving, dicarlo2007untangling, hung2005fast}, while the frontal lobe, associated with high-order cognition and semantic association, has limited direct contribution \cite{dicarlo2007untangling, fatourechi2007emg}. Effectively integrating high-level cognitive signals from the frontal lobe to assist reconstruction holds significant research value. Furthermore, existing metrics remain limited in providing a comprehensive assessment of the gap between reconstructed image and visual stimuli; generative models may synthesize images that are logically self-consistent but deviate from real perception. Thus, establishing a set of credibility evaluation systems combined with neuroscience standards is crucial for objectively evaluating brain decoding results.

To further explore the scope of application and limitations of the model, we selected several cases with poor reconstruction results for display, as shown in Fig. \ref{fig:failure}. When the original image is a ``bicycle," the reconstruction result only presents ``circular" outlines; when processing ``vertically arranged tools," the model simplifies and restores them into ``vertically oriented stick-like objects" of similar configuration. This suggests that cross-modal alignment could be further refined to better manage subtle feature nuances and visual representation ambiguity.


Finally, given the spatial resolution limits of non-invasive technology, invasive BCIs (iBCI)  remain essential for high-precision visual recovery through direct cortical recording\cite{zhao2023modulating}. In summary, CognitionCapturerPro enhances EEG-based visual decoding by optimizing alignment and fidelity, contributing to the development of more precise brain-computer interfaces.

\bibliographystyle{IEEEtran}
\bibliography{refs.bib}
\end{document}